**Sentiment Analysis in Poems in Misurata Sub-dialect**

A Sentiment Detection in an Arabic Sub-dialect

Azza Abugharsa, Ph. D

Department of Linguistics and Computer Science, Montclair State University, USA

**Abstract**

Over the recent decades, there has been a significant increase and development of resources for Arabic natural language processing. This includes the task of exploring Arabic Language Sentiment Analysis (ALSA) from Arabic utterances in both Modern Standard Arabic (MSA) and different Arabic dialects. This study focuses on detecting sentiment in poems written in Misurata Arabic sub-dialect spoken in Misurata, Libya. The tools used to detect sentiment from the dataset are Sklearn as well as Mazajak sentiment tool[1]. Logistic Regression, Random Forest, Naive Bayes (NB), and Support Vector Machines (SVM) classifiers are used with Sklearn, while the Convolutional Neural Network (CNN) is implemented with Mazajak. The results show that the traditional classifiers score a higher level of accuracy as compared to Mazajak which is built on an algorithm that includes deep learning techniques. More research is suggested to analyze Arabic sub-dialect poetry in order to investigate the aspects that contribute to sentiments in these multi-line texts; for example, the use of figurative language such as metaphors.

**Keywords:** Arabic, ALSA, sentiment, poems, Misurata

**1. Introduction**

With the increasing, easier access to multiple Internet channels and social media platforms in which opinions about different topics and products are expressed, natural language processing (NLP) tools and services have also shown a rapid improvement in order to investigate these structured and unstructured texts. Mining these utterances has led to the development of different applications and tasks. One of the most prominent tasks that have been intensively studied is classifying opinion sentiment in these utterances (Pang and Lee [54]; Hemmatian and Sohrabi [44]; Ghallab et al. [40]). Liu [48] defines sentiment analysis as the process by which sentiment in a given text is extracted and analyzed. The process includes classifying extracted opinions in these texts into either objective or subjective text (Ghallab et al. [40]). The subjective text can be classified into positive, negative, neutral, strongly negative, and strongly positive sentiments are also included in some classifications.

In this project, I investigate the sentiment in poems written in the Misurata Arabic sub-dialect spoken in Libya. Earlier research has focused on detecting sentiment from isolated utterances which are mostly taken from social media. To the best of my knowledge, no previous studies have tackled Arabic sentiment in dialectic poetry. Arabic poems include a sequence of sentences that are contextually and semantically connected to previous and subsequent sentences. All the sentences' content is based on the writer's main idea which reflects his feelings and thoughts.

To analyze the data, I use two different techniques to detect sentiment from poetry written in Misurata Arabic sub-dialect. The first technique is machine learning classifiers, and the second is a deep learning method. The purpose is to evaluate the performance of these methods on the given data. I use Sklearn to run the following classifiers: Logistic Regression, Random Forest, NB, and SVM. The second method is Mazajak which is designed on a CNN algorithm for the purpose of detecting sentiment in MSA and Arabic dialects. More discussion on these tools is provided in the Methodology section. The aim of this study is to answer the following question:

Is Mazajak more efficient for sentiment detection in Arabic sub-dialect poetry than other machine learning classifiers?

The following section presents a review of the related literature on sentiment analysis, with a description of the tools used to detect sentiment in Arabic. The next section discusses the process of sentiment classification in Arabic. After that, a description of the Misurata sub-dialect is provided. The section that follows talks about the focus of investigation in this study. After that, a description of the research methodology is presented, then the results are discussed and analyzed. Finally, the conclusion is provided.

## 2. Related work

Sentiment analysis, also known as opinion mining or the polarity classification of text, is one concept of NLP which involves exploring language aspects such as sentiment extraction, subjectivity analysis, affective computing, and emotional mining (Alsharif et al. [22]; Ahmed et al. [58]; Alsayat and Elmitwally [26]). These aspects are explored at three different scope levels of granularity; namely, the level of the document, the level of the sentence, and the level of aspect (Albayati et al. [11]). Boudad et al. [31] explain these three levels as follows:

- The document level: At this level, opinion extraction is performed on the whole document as a single entity. This can work with projects such as sentiment classification of a multi-line single piece of writing, e.g., poetry.

- The sentence level: It is the level at which the opinion polarities in the document sentences are considered individually for every single sentence. Unlike the document level, the sentiment prediction process in this level is challenging since it is context-dependent and in some cases needs to tackle sentences that are comparative and/or sarcastic.

- The aspect level: The purpose of this level is to meticulously perform two tasks: aspect detection with the attributes identified, and aspect sentiment labeling in which the aspects' sentiment orientation is defined.

Literature shows an intensive number of studies on sentiment analysis, most of which are conducted in English being the science language (Al-Sallab et al. [25]; Albayati et al. [11]). These studies explored different kinds of sentiment extraction. For example, while some studies provided an analysis of the sentiment of movies' reviews (Devi et al. [33]; Mamtesh and Mehla [50]), other projects have covered more topics and fields such as social media (Abu Kwaik et al. [6]; Al-Humoud [18]; Baali and Ghneim [29]; El-Halees [36]; Ghallab et al. [40], Heikal et al. [43]). Among these studies, several have focused on ALSA.

There are some challenges encountered when conducting ALSA. Some of these challenges can be related to factors such as the highly dialectical nature of Arabic, with a large variety of dialects and sub-dialects (Habash [41]; Darwish and Magdy [32]; Hammam et al. [42]; Alsiyat and Piao [27]). According to Boudad et al. [31], there are about thirty Arabic dialects spoken in different Arabic-speaking countries in the Middle East. These dialects are variant morphologically, phonologically, and lexically. In fact, in some cases, some dialects which are spoken in one Arabic country or region are hard to understand by native speakers of Arabic in another Arabic region especially if they are geographically distant.

Another factor is the rich morphological nature of Arabic with multiple synonyms for each word (Abu Farha and Magdy [32]; Elfaik and Nfaoui [35]). Arabic has a highly complex morphological system that can perform different tasks such as tokenization, lemmatization, part of speech tagging among others, Elfaik and Nfaoui [35] relate these challenges to aspects such as the ambiguity of Arabic dialects, the lack of contextual information, and the scarcity of words that express sentiment explicitly in Arabic texts.

Alsayat and Elmitwally [26] add one more potential factor which is the complexity of Arabic spelling, phonetics, vocabulary, semantics in addition to the challenge in understanding rhetorical questions and figurative ambiguity. The rhetorical aspect of Arabic belongs to the figurative level of the language frame. It includes the language of poetry in which rhetorical elements such as metaphors, personification, simile, and hyperboles are used. Nevertheless, the research domain of ALSA is growing and more findings and tools to detect sentiment in Arabic are generated.

*2.1 Tools to detect sentiment in Arabic*

There are different tools designed to detect sentiment in MSA in addition to other languages. One of these tools is SentiStrength, created by Theelwall et al. [60], which is originally designed in English but supports other languages such as Arabic. Likewise, the SemEval sentiment analysis tool, created by Rosenthal et al. [57] also includes sentiment detection in Arabic among other languages.

Deep learning techniques are very commonly used with ALSA. One example is Mazajak "your mood", which is an Arabic web-based tool developed by Abu Farha and Magdy [4,5]. This tool is based on deep learning algorithms designed on CNN. Mazajak, which is one of the tools used in this study, is applied to detect MSA as well as some Arabic dialects. Alsiyat and Piao [27] implemented Mazajak on their dataset to investigate the influence of metaphors on automatic ALSA. The findings show a significant drop in the tool's performance when it is used with sentences that include metaphors as compared to its performance with non-metaphorical sentences. In other words, their study shows that metaphors have a significant impact on ALSA algorithms and models. More details on Mazajak are provided in the Methodology section.

AraBert is another famous model that is used to conduct different NLP-related tasks. According to Antoun et al. [28], this model works to eliminate the need to employ word vectors when analyzing the given data. BERT is a language-specific model that proved to be efficient with many NLP tasks, especially with large corpus and datasets. The authors pre-trained BERT in the new model AraBERT to be used with Arabic aiming to make it generate similar results as the ones achieved with English. By comparing the performance of AraBERT with that of BERT, the results indicate achieving a high level of accuracy with AraBERT on most of the assigned Arabic NLP tasks including ALSA.

One more ALSA tool that also investigated different deep learning models based on CNN and The Long Short-Term Memory (LSTM) layer as a deep learning network is proposed by Al-Azani and El-Alfy [16]. The authors used word2vec based techniques to train neural models. One technique is the Continuous Bag of Words (CBOW), the other is the skip-gram. The results showed better performance with LSTM than CNN. The ensemble model has been implemented as well by Heikal et al. [43] by combining CNN and LSTM to detect sentiment in Arabic tweets. The accuracy score achieved was 65.05%.

Albayati et al. [11] introduce another tool to implement with such a rich morphologically structured language as Arabic. LSTM is implemented for training the model, and the word embedding is combined as a first hidden layer for features extraction. The model reached an accuracy level of 82%. This model is similar to Mazajak which uses LSTM and word embedding with a CNN as well. Galal et al. [39] Also used CNN to be very efficient with different NLP projects. The authors introduced GStem which is an algorithm in which word embeddings are used to group Arabic words that are similar; i.e., words that share the same roots-based on word vector distances. A CNN architecture is trained to tackle ALSA in different Arabic text documents.

Other more specific tools for ALSA are also designed. For example, Al-Samdi et al. [24] introduced an ALSA tool in which an SVM and Recurrent Neural Networks (RNNs) are used to detect hotel reviews. Also, Shoeb and Ahmed [58] used Naive Bayes (NB) and K-Nearest Neighbor (KNN) algorithms to extract sentiment polarity in Arabic. So far, studies of these dialects have been limited to major dialects such as Egyptian and Jordanian (Nabil et al. [52]; Almuqren et al. [20]; Al-Ayyoub et al. [24]; Alsayat and Elmitwally [26]; Alsiyat and Piao [27]. In one very recent study, Ali [12] uses different classification algorithms such as NB, KNN, SVM, in addition to Logistic Regression and Multinomial Naive Bayes to investigate COVID-19 related tweets and obtained a level of accuracy of 89.6%. It is my hope that the current study contributes to this body of research and provides a significant addition to the field of NLP in general and ALSA in specific.

*2.2 Sentiment classification in Arabic*

ALSA requires working with Arabic at different levels of the language's framework (Alsayat and Elmitwally [26]). This framework covers all aspects of Arabic which include: phonetics, morphology, syntax, lexicology, semantics, named entity recognition (NER), and figurative and rhetoric. Until now, each of these framework levels has been studied separately without considering a full hybrid parameter that combines all the language levels in one model architecture (Alsayat and Elmitwally [26]). According to Abdul-Mageed et al. [2], The stages of the classification process of these levels include data collection, data pre-processing, feature selection, classification techniques, and finally sentiment analysis.

By observing different projects conducted on ALSA, it is understood that sentiment in Arabic has three different classification methods. The first one is the lexicon-based approach in which the sentiment lexicon in different words and phrases that carry meanings of sentiment polarity is used to analyze attitudes, feelings, and opinions (Moussa et al. [51]). One of the projects in which the lexicon-based method is used is conducted by Al-Ayyoub et al. [14] who used lexicon derived from Arabic news articles to detect ALSA from Twitter data.

In their attempt to focus on the dialect variety of Arabic, Elmasry et al. [37] introduced a system that they called slang sentiment words and idiom lexicon, or (SSWIL). One more example is the study performed by Ibrahim et al. [45] on Arabic sayings and idiom phrases lexicon which are used to detect sentiment polarity in Arabic sentences by observing linguistically motivated features and the syntactic features that enhance the accuracy level of sentiment classification.

The second method of sentiment classification in Arabic is the machine learning classifier (Abdul-Mageed et al. [1]; Elarnaoty [34]; Almuqren et al. [20]; Baly et al. [30]) which involves using statistical machine learning algorithms in ALSA detection. Most of the work done on ALSA by the use of this method includes a combination of supervised machine learning algorithms such as decision trees, NB, SVM for large-size corpus, and KNN for smaller datasets.

The third method for sentiment classification in Arabic is the rule-based classifier (Oraby et al. [53]; Al-Radaideh and Al-Qudah [21]) in which ALSA is processed by using Arabic syntactic parsing rules. This method depends primarily on utilizing the polarity lexicon of three labels, namely positive, negative and neutral. According to Alsayat and Elmitwally [26], the classification accuracy of sentiment analysis is improved by using the mathematical tool rough set theory (RST) to reduce the feature vector's high dimension. In order to perform this kind of reduction, Al-Radaideh et al. [21] present the following four different reduction algorithms to enhance the overall accuracy level; FRAW, Genetic, Dynamic, and Exhaustive. The rule-based approach is applied to all levels of the Arabic language framework except for the phonetics and figurative ones (Alsayat and Elmitwally [26]). Hopefully, future work can cover these areas and derive useful findings.

*2.3 MSA and Misurata Arabic sub-dialect*

This section presents a brief discussion of the linguistic features that make Arabic one of the most challenging languages used in sentiment analysis works of research. First, a general discussion of MSA is provided followed by a discussion of the Misurata Arabic sub-dialect. Since the space here is not enough to go into details regarding standard Arabic and the Misurata dialect, only a brief description of these varieties is presented.

Arabic is among the top six official languages recognized by the United Nations. In addition, it is either the only official language or one of the official languages of 27 countries all located in the Middle East region. Arabic is spoken by more than 400 million people, most of whom are populated in these countries (Boudad et al. [31]).

Arabic is classified into three major categories: the first category is Classical Arabic which is the language of the Quran. The second category is MSA, and the third category is dialectical Arabic which is the informal variety of MSA spoken in everyday conversations in Arabic-speaking countries. Dialects vary from one country to another, and from one region to another. Furthermore, dialects branch out into different sub-dialects spoken in different regions within the same country. Since all these dialects stem out of the standard Arabic, they can generally be understood by everyone whose native language is Arabic.

The focus of the current study is to detect sentiment analysis from poems written in Misurata sub-dialect; this is one of the sub-dialects spoken in the country of Libya. Misurata, which is also referred to as the two-beach city and the sand city, is the third-largest city in the country, located on the Mediterranean coastal northern area of the country, about 130.49 miles east of the capital, Tripoli (Encyclopedia Britannica [61]).

Misurata dialect includes words that are mostly Arabic, among other borrowed words from Italian and Turkish languages as one result of these colonies that took place throughout history. As well, words from Tamazight (Berber) are included in this dialect due to the fact that the Berber is the native population of North Africa.

Arabic, as a standard variety, is essentially a VSO language, with the pattern SVO also recognized as the unmarked system. In the Misurata dialect, case endings are not attached to nouns and adjectives. Moreover, mood distinctions are not attached to verbs, and copular verbs are not used with sentences in the present tense (Al-Balushi [17]. One interesting attribute to notice about this dialect is that while duality is a significant feature that distinguishes standard Arabic from other languages, the dual affixes are absent in verbs, adjectives, and pronouns in this dialect. In other words, number in the Misurata dialect is indicated mainly by using singular and plural markers.

Regarding the sound system, the Misurata dialect has twenty-eight consonants and three main vowels. These vowels have two forms; a short form and a long-form with the vowel length being phonemic in the Misurata dialect as it is in Standard Arabic. In Misurata dialect, the consonant/vowel syllable structure includes the following types: [CV, CVV, CVC, CVCC, CCVV, CVVC, CCV, CCVVC, CCVCC, CCVC] (Elramli, 2012). As seen, the CC pattern is not favorable in the Misurata dialect (Maiteq [49]).

Morphology in the Misurata dialect is similar to that in standard Arabic. It has three different categories as derivational morphology which involves deriving new words from already existing ones, usually resulting in a word that belongs to a different part of speech. The second morphology category is inflectional morphology which is based on the process of conjugation that is very divergent in Arabic and is strongly related to gender and number of the agents. For example, the verb *write* has different inflectional affixes depending on the gender and number of the subject used with it. The third morphology category is agglutinative morphology in which different clitics or affixes can be added to a single word. As seen, Arabic is morphologically rich and a single shift in one of these suffixes results in a complete change in gender or number which could possibly result in the change of sentence meaning.

*2. 4 The significance of the current project*

The focus in the literature has been on detecting sentiment from isolated utterances which are mostly taken from social media (Abdulla et al. [3]; Abdul-Mageed et al. [2]; Refaee and Rieser [56]; Alayaba et al. [10]; Alsiyat and Piao [27]; Ghallab et al. [40]). The projects that have worked on Arabic poetry have focused on sentiment classification of standard Arabic old poetry (Alsharif et al. [22]) and MSA poetry (Ahmed et al. [58]).

Based on the literature I have studied, no previous work of ALSA has been performed on analyzing opinion sentiment in multi-line textual bodies as the case with dialectic Arabic poetry. Figure 1 which is introduced by Gallab et al. [40] shows the domains that have been under investigation when working on sentiment analysis. As Figure 1 indicates, most of the attention given to ALSA is directed towards social, political, and business domains, with less focus on other areas.

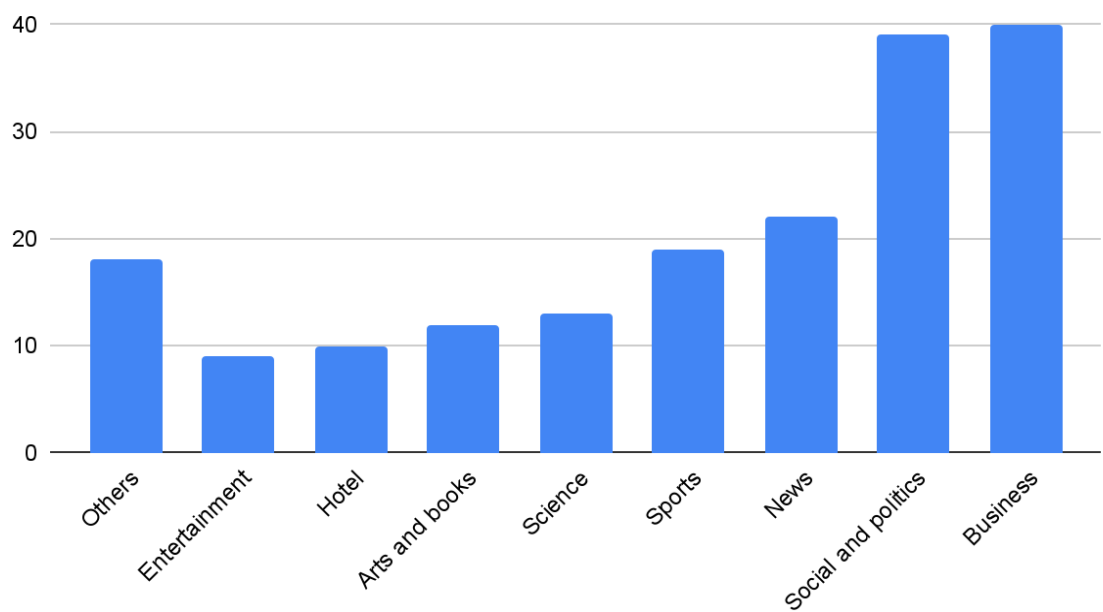

Figure 1. The most targeted domains in Arabic sentiment classification (Gallab et al. [40]).

Misurata Arabic sub-dialect poetry consists of one-versed poems in the form of separate lines which include sentences that are dependent upon sentences in other previous and subsequent lines. These attributes apply to Arabic poetry in general, both in MSA and in Arabic dialects. Since these sentences are contextually connected, the sentiment of the whole poem is reflected in each of these utterances. These features are not available in other kinds of data that have been investigated before. Therefore, such characteristics can constitute a new factor in Arabic sentiment classification, detection, and analysis.

Another gap in the literature is that all the previous ALSA projects have focused on extracting public opinions about a given topic or product. However, no work in Arabic has been done to computationally detect poets' opinions of multi-line textual bodies as is the case in poetry. Since no previous work is done on poetry that is written in any Arabic dialect or sub-dialect, one aim of this project is that it can lead to other subsequent projects that investigate different varieties of Arabic in different linguistic aspects.

One more point to add here is about the methods and tools used to investigate sentiment in Arabic. Many studies have used deep learning techniques in ALSA as well as other Sklearn tools such as decision trees, NB, and SVM (Al-Amrani [13]; Baali and Ghneim [29]; Elfaik and Nfaoui [35]; Zahidi et al. [63]). A comparison is made between different supervised learning classifiers vs. CNN and RNN and reported a significantly high-test accuracy level when using CNNs and RNNs. This study implements two approaches to ALSA: supervised learning classifiers and a CNN model. The purpose is to investigate the argument made in previous studies about deep learning networks being powerfully efficient machine learning techniques for ALSA as compared to other machine learning classifiers.

## 3. Methodology

In this section, I first introduce the dataset (3.1). Second, I discuss data cleaning and preprocessing (3.2). Then I provide a discussion about the methods (3.3). And finally, I discuss the classifier evaluation classifier (3.4).

*3.1 Dataset*

I use a corpus that I have built for the Arabic sub-dialect poems I have collected from the work of four poets. The dataset annotation is done manually by the researcher with the assistance and supervision of an expert in Arabic literature and poetry. The annotation process involves labeling the sentiment in the sentences into two classes; positive and negative. The expert initially classified the sentiment into two major labels: negative and non-negative; then classified the non-negative labels as positive. A neutral class can be used as well if the metaphors in the poems are considered, but this is beyond the scope of this project.

*3.2 Data preprocessing*

The data preprocessing step is required in order to reduce inconsistencies and make data easier to handle by normalizing the data into a coherent form. Table 1 shows the raw data before the preprocessing step.

| | |
|---|---|
| ذكرى بقت الأيام صارت اليوم احلام | **negative** |
| بيد ربنا الأقسام ماناش نفس التوب | **positive** |
| من البال مش ناسيك ديما نفكر فيك ولو غيرنا شاري... | **positive** |
| و الله لو بيديا تبقى معايا ليا | **positive** |
| لكن الدنيا هي اللي امفرقة الدروب | **positive** |
| يبان شوق القلب في نبضاته و يقولي ارجع علي مصراتة | **positive** |
| يا مجملها حلوة و يا مبهى طبايع هلها | **positive** |
| يا نار علي كان زرتها تعاود و تشتاقلها | **positive** |
| اللي زينها نحتار في صفاته | **positive** |
| شكرا ياغالي صحيت | **positive** |
| زعما كيف قدرت أتخون | **negative** |
| بالسية انتا اللي ابديت | **negative** |

**Table 1.** Raw data before the preprocessing step

To clean the data, a Python code is built to remove punctuation, hyphens, Arabic short vowels, diacritics, stop words (e.g., prepositions and conjunctions), empty lines, and elongation. In addition, normalization is performed in order to normalize letters into one unified form (e.g., replacing {أ, إ, آ} with {ا}, also {ئ} with {ي}, {ى} with {ا} and {ة} with {ه}). Since the dataset is built by me (the researcher), I was careful to avoid noise data, outliers, special characters, and empty lines. Table 2 shows the data after being preprocessed.

| | |
|---|---|
| ذكري بقت الايام صارت اليوم احلام | **negative** |
| ربنا الاقسام ماناش نفس التوب | **positive** |
| البال مش ناسيك ديما نفكر فيك غيرنا شاريك و اتف... | **positive** |
| و الله بيديا تبقي معايا ليا | **positive** |
| الدنيا اللي امفرقه الدروب | **positive** |
| يبان شوق القلب نبضاته و يقولي ارجع علي مصراته | **positive** |
| مجملها حلوه و مبهي طبايع هلها | **positive** |
| نار علي كان زرتها تعاود و تشتاقلها | **positive** |
| اللي زينها نحتار صفاته | **positive** |
| شكرا ياغالي صحيت | **positive** |
| زعما قدرت اتخون | **negative** |
| بالسيه انتا اللي ابديت | **negative** |

**Table 2.** The preprocessed data

*3.3 Methods*

This section describes the methods used in the study. First, I talk about the Python script (3.3.1), then I discuss the Mazajak ALSA tool (3.3.2).

*3.3.1 Python code*

A Python script is written to implement ALSA on my dataset by the use of four data mining classification algorithms: Logistic Regression classifier, Random Forest classifier, NB classifier, and SVM classifier. These algorithms have proven to be efficient with text classification and categorization, and have shown remarkable performance with other related studies (Shoukry [59]; Abdulla et al. [3]; Alsharif et al. [22]; Khasawneh et al. [47]; Alhumoud [19]; Nabil et al. [52]; Alshutayri et al. [23]; Ahmad et al. [7]; Ahmed et al. [8]; Hammam et al. [42]). These classifiers are mostly used with ALSA in both MSA as well as Arabic dialects. I aim to evaluate the performance of these classifiers on the Arabic sub-dialect in my data.

*3.3.2 Mazajak*

The second method I used is Mazajak, which is a deep learning tool designed to detect sentiment in MSA as well as major Arabic dialects. The model is designed on a CNN that works as a feature extractor based on learning the local patterns inside the phrases/sentences. Word2vec is used to create word embeddings that are fed into the CNN layer. After that, these embeddings are fed into a pooling layer where the features specific to sentiment are extracted. Next, these features are fed into an LSTM layer that works on the context and word order. Finally, the softmax layer provides the probability distribution over the output labeled classes. The whole process is depicted in Figure 2 presented by Abu Farha and Magdy (2019). As the designers of this model (Abu Farha and Magdy [4,5]) state, Mazajak is the second-largest word embedding set in Arabic after AraVec which was created by Soliman et al. [37].

**Figure 2.** The Majazak model architecture (Abu Farha and Magdy [4,5]).

Mazajak is used with my dataset as follows: the training data is fed sentence by sentence into the tool text input which is available for free access online. The text input is one of the four modes provided in the tool in which the system displays the polarity of the sentiment of the submitted text. Also, this mode provides a feedback option on the output sentiment so that this output can be corrected in case it is not accurate, and this is how the model learns. After that, the batch mode is used with the test data. In this process, the test data is submitted in a file and the model system returns an output file that contains the corresponding sentiment for each sentence in the test data.

*3.4 Evaluating the Performance of the Classifiers*

Since the dataset used in this project is relatively small, the 10-fold cross-validation (CV) technique is implemented for classifier evaluation. To evaluate the performance of the selected classifiers multiple times, the data is divided randomly into n-folds or parts. Each fold is held out once while the classifiers are trained on the remaining n-1 folds. The process is repeated until all the folds are used for training the classifiers. This technique helps avoid overfitting and gives the best results in learning the generic hypothesis.

**4. Findings**

By running the model codes, we get the results shown in Table 3 which displays the accuracy levels drawn from the classifiers used in the Python script.

|        | **Logistic Regression** | **Random Forest** | **Naive Bayes** | **SVM** |
|--------|-------------------------|-------------------|-----------------|---------|
| Python | 67 %                    | 68 %              | 69 %            | 68 %    |

**Table 3.** Accuracy scores in the four classifiers used in the Python Sklearn library

Table 3 shows that the four classifiers display different levels of test accuracy scores which are approximate to each other with NB classifier scores the highest rate.

As for Mazajak, the performance in the two sets of data is compared in order to detect the level of accuracy the model has achieved in learning from the training dataset. As shown in Table 4, there is no significant difference in the accuracy level achieved in the test set as compared to the training set.

| **Tool** | **Training Set** | **Test Set** |
|----------|------------------|--------------|
| Majazak  | 60.93 %          | 60.66 %      |

**Table 4.** The accuracy level of the training set and the test set used in Majazak

This, however, does not ignore the fact that the model managed to learn from the training set, even though this level of accuracy may not be significantly high for the data of this project. More research on this area is recommended in order to develop a model that can detect and analyze sentiment from all varieties of the Arabic language.

The overall performance of the two models is presented in Table 5. As seen, machine learning classifiers score a higher level of test accuracy than Mazajak which implements the CNN deep learning technique.

| **Tool**       | **Python** | **Majazak** |
|----------------|------------|-------------|
| Accuracy score | 68.0 %     | 60.66%      |

**Table 5.** Accuracy test scores of the three methods

This contradicts the findings in the literature of ALSA which are discussed by Ghallab et al. [40] who state that "Deep neural network has been successfully adopted to extract features. It has a big advantage over other ML methods." (Ghallab et al. [40], p. 15). This is confirmed by Albayati et al. [11] who state that significant achievements have been accomplished in sentiment analysis in the domain of deep learning in recent years.

Similarly, results in other studies on ALSA emphasize the efficiency of deep learning approaches in gaining high performance in the state-of-art deep learning models in the sentiment analysis domain as compared to traditional machine learning classifiers and methods (Ain et al. [9]; Al-Amrani [13]; Baali and Ghneim [29]; Heikal et al. [43]; Zahidi et al. [63]). One possible reason could be associated with the fact that there are no studies on ALSA in any Arabic sub-dialect, not to mention the lack of studies on the language of poetry which is rich with rhetoric and figurative speech.

There are two facts to consider here. The first one is that all the previous work conducted on Arabic poetry was on MSA, with no attention given to dialects and sub-dialects. The second one is that all the work was directed towards sentiment detection rather than sentiment analysis. These two facts emphasize the need for more investigation on the topic. Also, a tool such as Mazajak, though is specifically designed to detect sentiment from an input text in MSA and some major Arabic dialects, needs more training on Arabic sub-dialects, including detecting and analyzing sentiment from poetry.

**5. Conclusion**

This project investigates sentiment in poetry written in Misurata Arabic sub-dialect spoken in Libya. Although no previous work has been done on this kind of Arabic variety and on this kind of data, good results could be achieved and show that the traditional machine learning classifiers outperformed the deep learning model. Since this data is a collection of poems, it is undoubtedly rich with figurative language such as metaphors. In this regard, machine learning models need to be developed in order to detect the underlying cognitive mapping of metaphorical conceptualizations and other figurative processes. As a result, more work is recommended to explore the role the figurative level of the language can play in the process of sentiment and the detection of sentiment polarities in these kinds of contexts.

1.http://mazajak.inf.ed.ac.uk:8000/

**Acknowledgments**

I would like to express gratitude and appreciation to the four poets who provided me with their poems to be used in this project:

Salah Abaid

Alhusian Elasawadi

Mohamed Emmiama

Khalid Iqsheerah

I also want to thank Mrs. Amna Omar Ben Hamaida, an expert in Arabic literature, rhetoric, and morphology for her great assistance in classifying and detecting the sentiment in these poems.